\documentclass[letterpaper, 10 pt, journal, twoside]{IEEEtran}

\IEEEoverridecommandlockouts                              % This command is only needed if 
\usepackage{graphics} % for pdf, bitmapped graphics files
\usepackage{epsfig} % for postscript graphics files
\usepackage{amsmath} % assumes amsmath package installed
\usepackage{soul}
\usepackage{xcolor}
\usepackage{fancyhdr}

\usepackage{algorithm}% http://ctan.org/pkg/algorithm
\usepackage{amsmath}
\DeclareMathOperator*{\argmax}{arg\,max}

\usepackage{algpseudocode}% http://ctan.org/pkg/algorithmicx
\usepackage{array}
\usepackage{mdwmath}
\usepackage{mdwtab}
\usepackage{eqparbox}
\usepackage{amssymb}
\usepackage{subfig}
\usepackage{gensymb}

\title{
Constrained-Space Optimization and Reinforcement Learning for Complex Tasks
}

\author{Ya-Yen Tsai\IEEEauthorrefmark{2}, Bo Xiao\IEEEauthorrefmark{2}~\IEEEmembership{Member,~IEEE}, Edward Johns and % <-this % stops a space
 Guang-Zhong Yang,~\IEEEmembership{Fellow,~IEEE}
\vspace{-5mm}
 \thanks{Manuscript received: September, 9, 2019; Revised December, 1, 2019; Accepted December, 23, 2019.}
 \thanks{This paper was recommended for publication by Editor Pietro Valdastri upon evaluation of the Associate Editor and Reviewers' comments. This work was supported by Engineering and Physical Sciences Research Council (EPSRC) under Grant (EP/L020688/1).}
\thanks{\IEEEauthorrefmark{2} indicates equal contribution. Y.-Y. Tsai, B. Xiao and G.-Z. Yang are with the Hamlyn Centre for Robotic Surgery and E. Johns is with the Robot Learning Lab, Imperial College London, SW7 2AZ, London, UK (e-mail: \{y.tsai17, b.xiao, g.z.yang, e.johns\}@imperial.ac.uk). G.-Z. Yang is also with the Institute of Medical Robotics, Shanghai Jiao Tong University, China.}
\thanks{Digital Object Identifier (DOI): 10.1109/LRA.2020.2965392}}
\newcommand{\changefont}{%
    \fontsize{7}{10}\selectfont
}
\fancypagestyle{firststyle}
{

\fancyhead[L]{\changefont \MakeUppercase{ IEEE Robotics and Automation Letters. Preprint Version. Accepted December, 2019}}
\fancyhead[R]{\changefont \thepage}

   \fancyfoot[C]{
   \vspace{-5mm}
   \changefont $\textcopyright$ 2020 IEEE. Personal use of this material is permitted.  Permission from IEEE must be obtained for all other uses, in any current or future media, including reprinting/republishing this material for advertising or promotional purposes, creating new collective works, for resale or redistribution to servers or lists, or reuse of any copyrighted component of this work in other works.}
}

\begin{document}

\maketitle

\thispagestyle{firststyle}

% \thispagestyle{empty}
% \pagestyle{empty}

% Paper headers 
\markboth{IEEE Robotics and Automation Letters. Preprint Version. Accepted December, 2019} {Tsai \MakeLowercase{\textit{et al.}}: Constrained-Space Optimization and Reinforcement Learning for Complex Tasks}
% Use only for final RAL version

%%%%%%%%%%%%%%%%%%%%%%%%%%%%%%%%%%%%%%%%%%%%%%%%%%%%%%%%%%%%%%%%%%%%%%
\begin{abstract}

% This electronic document is a `live' template. The various components of your paper [title, text, heads, etc.] are already defined on the style sheet, as illustrated by the portions given in this document.
Learning from Demonstration is increasingly used for transferring operator manipulation skills to robots. In practice, it is important to cater for limited data and imperfect human demonstrations, as well as underlying safety constraints. This paper presents a constrained-space optimization and reinforcement learning scheme for managing complex tasks. Through interactions within the constrained space,  the reinforcement learning agent is trained to optimize the manipulation skills according to a defined reward function. After learning, the optimal policy is derived from the well-trained reinforcement learning  agent, which is then implemented to guide the robot to conduct tasks that are similar to the experts' demonstrations. The effectiveness of the proposed method is verified with a robotic suturing task, demonstrating that the learned policy outperformed the experts' demonstrations in terms of the smoothness of the joint motion and end-effector trajectories, as well as the overall task completion time.

\end{abstract}

\begin{IEEEkeywords}
Medical robotics, Learn from Demonstration (LfD), Reinforcement Learning (RL), Robot learning, Robotic suturing.
\end{IEEEkeywords}
%%%%%%%%%%%%%%%%%%%%%%%%%%%%%%%%%%%%%%%%%%%%%%%%%%%%%%%%%%%%%%%%%%%%%%%%%%%%%%%%
\section{INTRODUCTION}
\IEEEPARstart{R}{obot} learning has facilitated the programming of a robot using Learning from Demonstrations (LfD) and therefore has gained its popularity in the past decades \cite{atkeson1997robot, argall2009survey}. One objective of robot learning is to endow the robot with the ability to learn tasks through its own observations. By performing a task by a teacher, a robot can learn, and thus replicate what has been demonstrated. Compared to approaches that learn from scratch, higher learning efficiency can be realized through this approach. Provision of human demonstration transfers the knowledge/skills from the teacher to the learner, avoiding the robot to relearn those already acquired by and can be transferred from the demonstrator.

 However, providing a high quality demonstration can be expensive and may not always be possible, not to mention the necessity of large number of demonstrations required to cover enough state-action pairs for policy learning. LfD implies that the robot policy may inherit the underlying motion characteristics of the teacher and as such, the poor demonstrations may affect the final performance of the robot. Besides, due to the difference in the morphological structure between the teacher and the learner, direct replication of the demonstrated task by the robot may not be efficient and sometimes not feasible. Hence, with the advances in robot learning, many researches have moved from simply teaching the robot, to exceeding the performance of human~\cite{van2010superhuman}.

Improving the learning performance from imperfect demonstration can be formulated as an optimization problem. Reinforcement Learning (RL) can be a good candidate to serve for this purpose. By defining a reward function, the learning agent can find the optimal policy such that it maximizes the cumulative rewards received. However, a typical problem is to find a good balance between exploitation and exploration during the learning process. A good exploitation of data may not be possible without sufficient exploration while too much exploration may degrade the learning efficiency. Furthermore, for a surgical robot, as an example, improper or too wild exploration may raise in safety concerns. Therefore, in this context, we propose a framework that can learn from sub-optimal demonstrations using RL and performs policy learning within a bounded discretized space, which is  constrained by variance of the human demonstrated trajectory. This is shown to shorten the exploration time and enhance the policy learning performance. 

\begin{figure}
    \centering
    \includegraphics[width=0.85\linewidth]{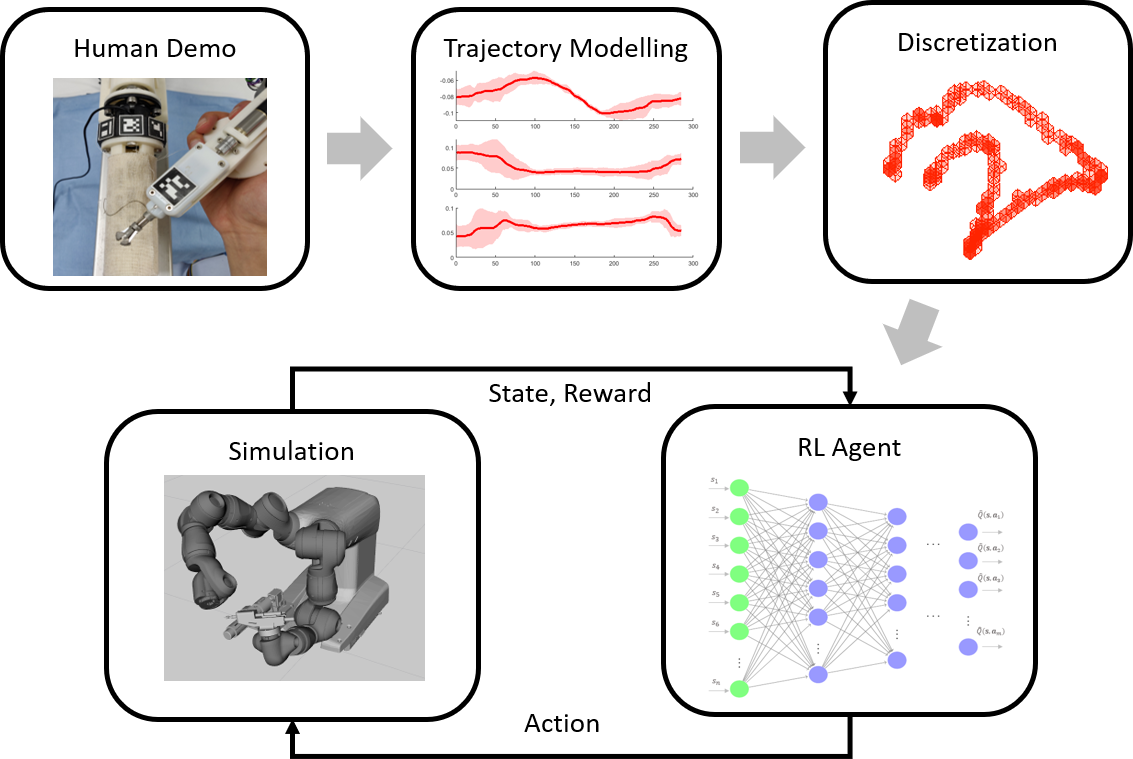}
    \caption{An overview of the proposed framework illustring how trajectory modelling is discretized and learned through an RL agent.}
    \label{fig:overview}
    % \vspace{-3mm}
\end{figure}

\section{RELATED WORK}
\textcolor{black}{In robot learning, LfD has been used in an intuitive way to transfer human knowledge to a robot. For example, Mueller et. al~\cite{mueller2018robust} proposed an approach to repair errors in acquired skills through additional demonstrations while Osa et. al~\cite{osa2017guiding} incorporated human demonstrated trajectories in addition to hand-crafted constraints to avoid the need for motion planning from scratch. In this way, not only the learning process through LfD can be made more efficient, but also the robot has the potential to perform a task with quality that is on par with that of an expert.}

% LfD example
A general technique for policy derivation is to train a classifier or regressor on a set of state-action pairs obtained from the demonstrated examples~\cite{argall2009survey}. Taking the state as training input and the action as the output label, this approach aims to approximate a mapping function directly from a state to an action. Gaussian Mixture Models and Gaussian Mixture Regression (GMM/GMR) derive the approximation by first encoding each motion primitive using GMM then applying GMR to concatenate models and generate smooth trajectories~\cite{reiley2010motion, lin2012learning}. Dynamic Movement Primitives (DMP) proposed by Ijspeert et al.~\cite{ijspeert2002movement}, on the other hand, use a set of differential equations and a nonlinear term to model each motion primitive as a nonlinear dynamical system. Multiple equations describing motion primitives are combined to form a smooth trajectory~\cite{pastor2009learning, ijspeert2013dynamical}. Other work used variations of local regression based on k-Nearest Neighbors, locally weighted regression~\cite{atkeson1991using}, receptive field weighted regression~\cite{schaal1997receptive} and locally weighted projection regression~\cite{vijayakumar2000locally} to map nonlinear high dimensional problems. These approximators attempt to model and learn the demonstrator's behaviour to a state from the demonstrations, so they could carry out an action similar to that of the teacher. However, to fully capture a demonstrator's motion characteristics, an approximator would have to rely heavily on multiple critical demonstrations, which in practice, are hard to obtain. 

Various application problems in robotics can be formed and solved in the realm of RL\textcolor{black}{~\cite{kober2013reinforcement, rusu2016sim}}. Through iterative interaction with an environment, different states, actions and the associated rewards can be explored, and hence, an optimal policy can be learned by solving the corresponding Markov Decision Process (MDP)~\cite{sutton1998introduction}. Without the need for excessive demonstrations, RL has shown its potential in finding an optimal policy in many applications~\cite{silver2016mastering, ebrahimi2017gradient, gao2018reinforcement, hester2018deep}. Vecerik et. al~\cite{vevcerik2017leveraging} leveraged this concept and proposed a Deep Deterministic Policy Gradient from Demonstrations (DDPGfD). The framework incorporated the experience of demonstrations to its replay buffer and prioritized it such that the more important transitions are sampled more frequently. Similar work proposed by Nair et. al~\cite{nair2018overcoming}, also included the experience of demonstrations in replay buffer, but additionally, introduced an auxiliary loss and Q-filter to enhance the learning performance. An introduction of RL avoids the need for large scale demonstrations, but to some extent, it still requires sufficient high quality demonstrations in order to find the optimal policy.

Instead of using the demonstrations to initialize the policy, Kim et. al~\cite{kim2013learning}, on the other hand, proposed Approximate Policy Iteration with Demonstration (APID) using the information from a few and/or sub-optimal demonstrations as imposed constraints to an optimization problem, making it less prone to noisy demonstrations. Kang et. al~\cite{kang2018policy} proposed a Policy Optimization from Demonstration (POfD), which utilized the scarce and imperfect demonstrations to limit the exploration region to be around that of the expert policy. By including the demonstration-guided exploration term to its learning objective, the author has shown that a better policy could be achieved. Considering the merits of RL and taking the idea of imposed constraints\textcolor{black}{~\cite{osa2017guiding}} from the scarce and sub-optimal demonstrations, we proposed in this paper a framework based on Deep Reinforcement Learning (DRL) approach to find an optimal policy from these demonstrations without losing the safety guarantees. The contributions of the proposed DRL approach are \textcolor{black}{summarized} as follows:

\begin{enumerate}
    \item The proposed DRL framework learns an optimal skill for a sewing task from scarce and imperfect human demonstrations 
    \item The statistical distributions of the multiple human demonstrations are utilized to construct the confined space that satisfies the safety requirement for optimization.
    \item The RL agent is trained within the constrained space to improve the learning performance and efficiency.
\end{enumerate}

% the structure of the paper
The organization of the rest of the paper is as follows. In Section III and IV, the preliminaries and the methodology of the proposed framework are introduced. The experimental setup, the validation of the method and the discussion are then presented in Section V followed by the conclusions and future works in Section VI.

\section{PRELIMINARIES}
In this section, the preliminaries of MDPs, Q-learning and deep Q-learning will be presented.
\subsection{Markov Decision Processes}
\textcolor{black}{A} MDP is a mathematical framework based on state set $\mathcal{S}$ associated with finite action set $\mathcal{A}$. If and only if state $\mathbf{s}$ in $\mathcal{S}$ captures all the relative information from history, state $\mathbf{s}$ is Markovian. The transition function between two Markov states can be defined as: $f: \mathcal{S}\times \mathcal{A} \rightarrow \mathcal{S}$. In addition, an Markov Reward Process (MRP) is defined as the tuple $<\mathcal{S}, f, r, \gamma>$, where $r$ is a scalar reward function and $\gamma$ is a discount factor. The scalar reward function is defined as $r: \mathcal{S}\times \mathcal{A}\times\mathcal{S} \rightarrow \mathbb{R}$. Parameter $\gamma \in [0, 1]$ determines how one values the importance of current reward and future rewards. Besides, $\gamma$ also stabilizes the total return in infinite time-step case. An MRP with control input/action is defined as an MDP: $<\mathcal{S}, \mathcal{A}, f, r, \gamma>$. The objective of an RL agent is to find the optimal policy that maximizes the accumulated rewards (return) $R$. Considering the observed states $\mathbf{s} \in \mathcal{S}$ and admissible control input $\mathbf{a} \in \mathcal{A}$, the overall learning problem can be formed into MDPs. To evaluate the value function of the policy, the RL agent will try to solve the MDP. When the value function is obtained, the RL agent can make the optimal decisions according to it, and thus to fulfill the decision-making objective.

\subsection{Q-learning}
In RL, the behavior of the RL agent is determined by the policy defined as $\pi(\mathbf{a}\vert\mathbf{s}) = P(\mathbf{a}_\omega = \mathbf{a}\vert \mathbf{s}_\omega = \mathbf{s})$. The policy $\pi(\mathbf{a}\vert\mathbf{s})$ represents the probability distribution of the action picking $\mathbf{a}_\omega$ according to the observed system state $\mathbf{s}_\omega$. $r_{\omega}$ is the reward obtained after yielding the action during the $\omega$-th time-step. The RL agent takes the observed state $\mathbf{s}_\omega$ as input and gives out the action $\mathbf{a}_\omega$, at the same time, receives the reward $r_\omega$ at time-step $\omega$. Associated with the reward function, the state-action value $Q(\mathbf{s}_\omega, \mathbf{a}_\omega)$ in RL is defined as the expectation of return $R_\omega$ which starting with state $\mathbf{s}_\omega$ and action $\mathbf{a}_\omega$. The state-action value function under policy $\pi$ can be calculated as follows:
\begin{align}
	Q^{\pi}(\mathbf{s}, \mathbf{a}) & = \mathbb{E}_{\pi}[R_\omega\vert \mathbf{s}_\omega = \mathbf{s}, \mathbf{a}_\omega = \mathbf{a}] \notag \\
	& = \mathbb{E}_{\pi}[r_\omega + \gamma r_{\omega+1} + \gamma^2 r_{\omega+2} + \ldots\vert \mathbf{s}_\omega = \mathbf{s}, \mathbf{a}_\omega = \mathbf{a}] \notag \\
	& = \mathbb{E}_{\pi} [\sum_{k = 0}^{\infty}\gamma^kr_{\omega+k}\vert\mathbf{s}_\omega = \mathbf{s}, \mathbf{a}_\omega = \mathbf{a}]   
\end{align}

and the state-action value function can be also rewritten as the Bellman Equation (BE): 
\begin{align}
	Q^{\pi}(\mathbf{s}_\omega, \mathbf{a}_\omega) = \mathbb{E}_{\pi}[r_\omega + \gamma \mathbb{E}_{\pi}[Q^{\pi}(\mathbf{s}_{\omega+1}, \mathbf{a}_{\omega+1})]].
\end{align}

The goal of Q-learning is to find the optimal decision-making policy $\pi$, which maximizes the state-action value function. The optimal state-action value function can be defined as:
\begin{align}
		Q^{*}(\mathbf{s}, \mathbf{a}) &= \max \limits_{\pi}\mathbb{E}_{\pi} [r_\omega + \gamma r_{\omega+1} + \gamma^2 r_{\omega+2}  \notag \\
		&+ \ldots \vert \mathbf{s}_\omega = \mathbf{s}, \mathbf{a}_\omega = \mathbf{a}].
\end{align}

To maximize the return obtained by the RL agent, the Q-learning algorithm updates the state-action values according to the reward function in an off-policy style as:
\begin{align}\label{eq:Q-learning}
{Q}(\mathbf{s}_\omega, \mathbf{a}_\omega) & \leftarrow  (1-\alpha)Q(\mathbf{s}_\omega, \mathbf{a}_\omega) \notag \\
& + 
\alpha(r_\omega + \gamma\max\limits_{\mathbf{a}_k}\{{Q}(\mathbf{s}_{\omega+1}, \mathbf{a}_k)\}),
\end{align}
where $\alpha \in (0, 1]$ is the learning rate, which determines the learning speed of Q-learning.

To ensure that Q-learning explores extensively without diverging, the adaptive $\epsilon$-greedy policy is adopted in this paper. To change the value of the exploration rate $\epsilon$ in an adaptive way, we start from large value of $\epsilon$ and then reduce it gradually after episodes being completed. 
\begin{align}
	\mathbf{a}_\omega = \left\{  
	\begin{array}{lr}  
	 \argmax \limits_{\mathbf{a}_k} \{{Q}(\mathbf{s}_{\omega}, \mathbf{a}_k)\}, \ \text{with probability:} \ 1 - \epsilon,  \\
	 \text{random}\ \mathbf{a}_k, \ \text{with probability:} \ \epsilon.
	\end{array}  
	\right.  
\end{align}

To summarize, the Q-learning algorithm can be viewed in Algorithm \ref{afql}.
\begin{algorithm}
	\caption{Algorithm for Q-learning}\label{afql}	
	\begin{algorithmic}[1]
		\State Set the state-action values randomly
		\For{Every episode}
		\State Initialize state $\mathbf{s}_0$
		\For{Current timestep $\omega$}
		\State Pick action $\mathbf{a}_\omega$ through $\epsilon$-greedy policy and observe the next state $\mathbf{s}_{\omega+1}$
		\State Update ${Q}(\mathbf{s}_\omega, \mathbf{a}_\omega) \leftarrow (1-\alpha)Q(\mathbf{s}_\omega, \mathbf{a}_\omega) + 
		\alpha(r_\omega + \gamma\max\limits_{\mathbf{a}_k}\{{Q}(\mathbf{s}_{\omega+1}, \mathbf{a}_k)\})$
		\State $\mathbf{s}_\omega \leftarrow \mathbf{s}_{\omega+1}$
		\EndFor 
		\EndFor 
	\end{algorithmic}
\end{algorithm}

\subsection{Deep Q-learning}
When the state set $\mathcal{S}$ and action set $\mathcal{A}$ are too large or continuous, calculation of the exact state-action value function $Q(\mathbf{s}, \mathbf{a})$ becomes difficult and may be impossible in many cases. One alternative is to use a function approximator to replace the exact state-action value function. As an universal approximator, neural networks with nonlinear activation functions can be used for the Q-network to approximate ${Q}(\mathbf{s}, \mathbf{a})$ \cite{hornik1990uaf}. When the state-action value function is approximated, RL now becomes approximate reinforcement learning.

 In the above algorithm, the observed state $\mathbf{s}$ is the input of the deep neural network while the output $\hat{Q}(\mathbf{s}, \mathbf{a}_k)$ approximates the exact state-action value $Q(\mathbf{s}, \mathbf{a}_k)$, $k =1$, $2$, $\ldots$, $m$. By choosing the action according to the largest state-action value $\hat{Q}(\mathbf{s}, \mathbf{a}_k)$, the RL agent is able to make the optimal decision. 
 
 To simplify the notations, the set of all parameters (including all the weights and biases) in the deep Q-network will be written as $\theta^D$ in the rest of this paper. The deep Q-network implicitly determines the policy, in which the action $\mathbf{a}_k$ is picked according to the observation $\mathbf{s}_\omega$ at time-step $\omega$. The chosen action drives the current state $\mathbf{s}_\omega$ to the next state $\mathbf{s}_{\omega+1}$, and the reward $r_\omega$ will be received from the environment according to the reward function. Then the target network takes the state $\mathbf{s}_{\omega+1}$ and the control input set $\mathcal{A}$ as the input to calculate the output $\max\limits_{\mathbf{a}_k}\{\hat{Q}(\mathbf{s}_{\omega+1}, \mathbf{a}_\omega\vert \theta^{D})\}$ to estimate largest state-action value for the next state $\mathbf{s}_{\omega+1}$. When $\max\limits_{\mathbf{a}_k}\{\hat{Q}(\mathbf{s}_{\omega+1}, \mathbf{a}_\omega\vert \theta^{D})\}$ and $r_\omega$ are available, the state-action value $\hat{Q}(\mathbf{s}_\omega, \mathbf{a}_\omega\vert \theta^{D})$ of the current state-action pair can be updated according to the target network and obtained reward. Considering the importance of experience replay during the training of the RL agent~\cite{mnih2015human}, the state transitions associated with rewards are store in the experience buffer as the tuple $<\mathbf{s}_i, \mathbf{a}_i, \mathbf{s}_{i}', r_i>$. $\mathbf{s}_i$ is the current state, $\mathbf{a}_i$ is the chosen action according to the current state, $\mathbf{s}_{i}'$ is the next state and $r_i$ is the reward received. The algorithm for updating of the weights of the deep Q-network is presented in detail in Algorithm \ref{euclid}.
\begin{algorithm}
	\caption{Update of the parameters in the deep Q-network}
	\label{euclid}	
	\begin{algorithmic}[1]
%		\State Choose the structure for the Q-network
		\State Initialize the parameters in the deep Q-network  $\theta^{D}$: $$ \theta^{D} \leftarrow \theta^{D_0}$$
		\State Create the experience buffer (EB) 
		\State Define the constants in the training: Num\_Episode, Num\_Time\_Step,  Num\_Replay\_Times
		\For{\texttt{{Episode = 1:Num\_Episode}}
		}
		\State Choose the initial state $\mathbf{s}$ randomly

		\For{\texttt{{j\_time = 1:Num\_Time\_Step}}
		}
		\State Generate action $\mathbf{a}_\omega$ through policy $\pi$ determined by the current deep Q-network
		\State Transmit to the next state $\mathbf{s}_{\omega+1}$ and observe the reward $r_\omega = r(\mathbf{s}_{\omega}, \mathbf{a}_\omega, \mathbf{s}_{\omega+1})$ 
		\State Store the tuple $<\mathbf{s}_\omega, \mathbf{a}_\omega, \mathbf{s}_{\omega+1}, r_\omega>$ in the EB
		\State $\mathbf{s}_{\omega} \leftarrow \mathbf{s}_{\omega+1}$
%		\EndFor 
%		\State \textbf{Learning Stage:} 
%		\For{\texttt{{e\_time = 1:Num\_Replay\_Times}}
%		}
		\State Randomly sample a mini-batch of $N$ transition tuples $<\mathbf{s}_i, \mathbf{a}_i, \mathbf{s}_{i}', r_i>$, $i=1, 2, \ldots, N$ from the EB
		\State Obtain the Q-target as:
		$$\bar{Q}_i = r_i + \gamma\max\limits_{\mathbf{a}_k}\{{Q}(\mathbf{s}_{i}', \mathbf{a}_k|\theta^{D})\}$$
		\State Calculate the cost function: $$J(\theta^D) =  \frac{1}{2N}\sum_{i=1}^{N}(\bar{Q}_i - {Q}(\mathbf{s}_i, \mathbf{a}_i)|\theta^D)^2$$
		\State Calculate the mini-batch gradient of $\theta^D$: $$\nabla_{\theta^D} J(\theta^D) = -\frac{1}{N}\sum_{i=1}^{N}(\bar{Q}_i - {Q}(\mathbf{x}_i, \mathbf{a}_i)|\theta^D)\frac{\partial {Q}(\mathbf{s}_i, \mathbf{a}_i|\theta^D)}{\partial \theta^D}$$
		\State Update the weights of the deep Q-network through gradient descent: $$\theta^D \leftarrow \theta^D - \alpha^D\nabla_{\theta^D} J(\theta^D),$$ 
		\EndFor
		\EndFor
	\end{algorithmic}
\end{algorithm}
\section{METHODOLOGY}
Based on the DRL algorithm presented above, we propose a framework that extends from the existing LfD approach to enhance the overall robot performance. From a small set of human demonstrations, the modelling is done at the trajectory level to capture the underlying characteristics and distributions of human motions. To further refine the skill learned from  human demonstrations, the DRL approach mentioned above is adopted for the optimization. Considering the safety requirement on the robot, the mean and the variance of the trajectory serve as the constraints to the unexplored state space within which the optimal trajectory will be found. The RL agent is then trained and the corresponding policy is optimized through interaction with the physical simulator based on the pre-modelled trajectories from the previous step. 

\begin{figure}
    \centering
    \includegraphics[width=0.9\linewidth]{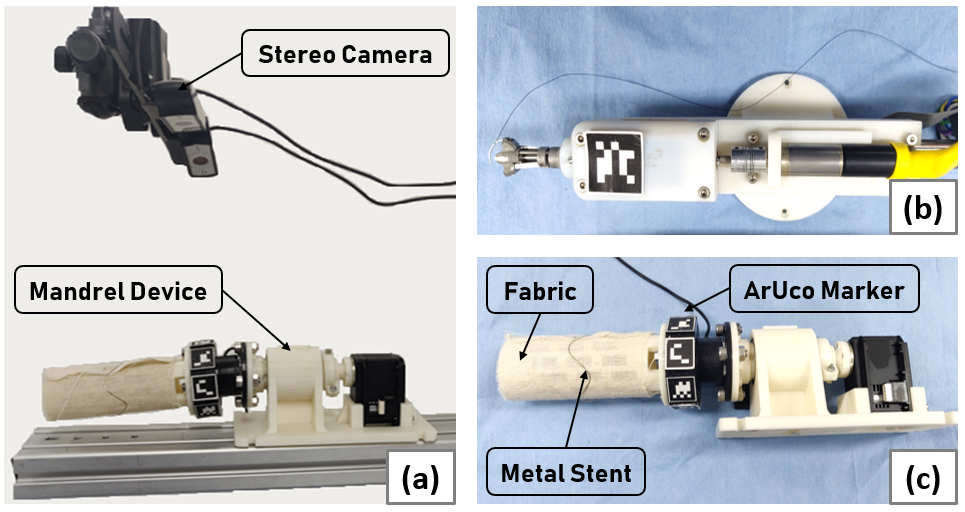}
    \caption{(a) The setup of the human demonstration, that uses the stereo camera to track the markers attach to a device (b) The hand held suturing device and (c) The mandrel device.}
    \label{fig:setup}
\end{figure}

\subsection{Human Demonstration and Problem Formulation}
For this paper, the setup of human demonstration emulates a robot manufacturing task for stent graft~\cite{huang2017vision,tsai2019transfer}, which consists of three components: a stereo camera, a suturing device~\cite{hu2019design} and a mandrel device as illustrated in Fig. \ref{fig:setup}. The stereo camera is to capture the motions of the suturing device and the mandrel device during the demonstration. The suturing device is a hand held device, that facilitates single-handed stitching by passing its needle between two ends at the tip. The mandrel device consists of a cylindrical supporting structure, covered by a piece of fabric and a metal stent, with slot windows to allow for needle piercing. 

\begin{figure}
    \centering
    \subfloat[\scriptsize{Continuous 3D trajectory}]
    {\includegraphics[width=4cm]{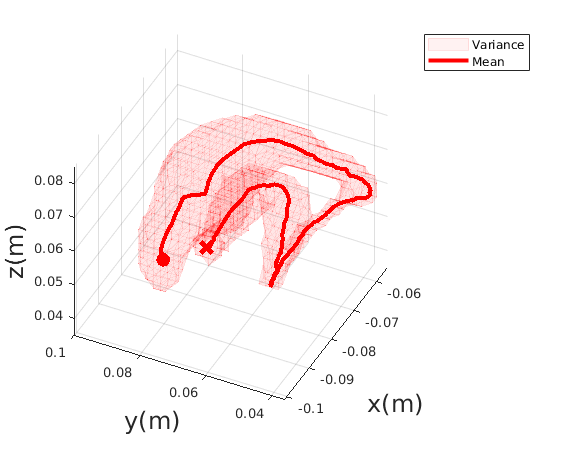}}
    \subfloat[\scriptsize{Discretized 3D trajectory}]
    {\includegraphics[width=4.2cm]{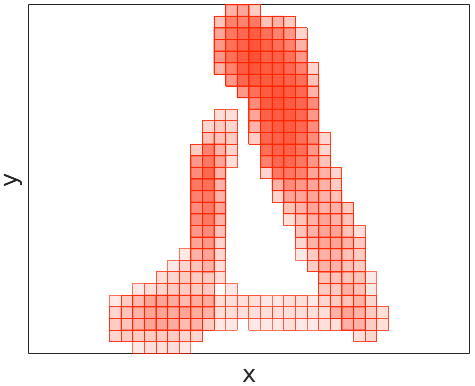}}
    \caption{Comparison of (a) continuous and the (b) (top view of the) discretized task space computed from the mean and the variance of the demonstrated trajectories. The solid circle and the cross represent the starting and the end points respectively.}
    \label{fig:3d_mean_std}
    % \vspace{-3mm}
\end{figure}

The task is to teach a robot to stitch on a specified slot such that it binds the stent and the fabric tightly. A human demonstration manipulates the suturing device around the desire location while its motion is tracked using an ArUco marker and the stereo camera. The stitching process consists of three motion primitives, approach the stitch slot, pierce in and pass the needle, and return back to the initial pose. The overall trajectory is illustrated in Fig.~\ref{fig:3d_mean_std}.

\begin{figure}
    \centering
    \includegraphics[width=0.9\linewidth]{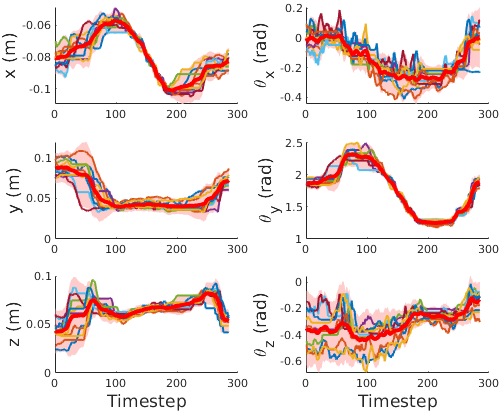}
    \caption{10 aligned demonstrated trajectories from the 6 DoF. The mean trajectory is shown using the red line and the variance is labelled using the shaded region.}
    \label{fig:6d_mean_std}
    % \vspace{-5mm}
\end{figure}

The recording of demonstrations was done at a fixed sampling rate and it captured the movements of the suturing device and the mandrel device. As the mandrel is subject to movements under the camera frame, by capturing its motion allows finding the relative pose between the two devices. 

Each demonstration records the trajectory, $\mathcal{S} = \{\gamma_m, \gamma_s\}$, which forms from the poses of the mandrel and the suturing device under the camera frame. The action performs a stitch on the same location of the mandrel over time. $\gamma\in\mathbb{R}^{6}$ is a 6 DoF trajectory, which consists of the translation component, $T = [x, y, z]$ and the rotation component, $R = [\theta x, \theta y, \theta z]$, in Euler-Rodrigues representation. A collection of human demonstrations is recorded and used to capture the underlying characteristics of demonstrator's hand motions.

\subsection{Trajectory Modelling}
Multiple trajectories are \textcolor{black}{pre-processed} before modelling. The first part of the pre-processing is to filter the noises and sudden hand movements through smoothing the trajectories. Then, the trajectory of the suturing device is transformed to the mandrel frame. A suturing device's trajectory is oriented around the mandrel and is subject to changes if the pose of the mandrel changes in the camera frame. Therefore, for modelling, the trajectory of interest is the device' trajectory with respect to that of the mandrel device. This forms a 6 DoF trajectory $\gamma_{ms}\in\mathbb{R}^{6}$ consists of the translation component, $T_{ms} = [x_{ms}, y_{ms}, z_{ms}]$ and the rotation component, $R_{ms} = [\theta x_{ms}, \theta y_{ms}, \theta z_{ms}]$, in Euler-Rodrigues representation for each demonstration. 

Trajectory modelling is done temporally and the purpose is to find the mean and variance throughout all the demonstrated data. Multiple trajectories possess different temporal length resulted from variations in device's manipulation. They are aligned temporally using Dynamic Time Warping (DTW) and all the DoF. DTW is a commonly used distance-based method to compute similarity among data set and perform temporal sequences alignment such that the distance score is minimized. After which, the mean and standard deviation is found at each temporal point throughout the aligned trajectories. This forms a 3D space bounded by a standard deviation and within which the optimal trajectory will be found.

\subsection{Trajectory Optimization}
In practice, motion redundancy in a human demonstrated trajectory and the morphological differences between the demonstrator as well as the learner make the demonstration sub-optimal and prevent it from direct replay on a robot. Motion redundancy is unnecessary movement from tracking errors or human factors that can result in sudden unexpected motion or longer trajectory execution duration. 
The morphological difference, on the other hand, is the imperfect mapping between the performer and the learner, hence without optimization, the demonstrated trajectory may result in unsmoothed joint movements in a robot even if the demonstration was perfect, which may raise safety and collision concern.  

\begin{figure}
    \centering
    \includegraphics[width=0.7\linewidth]{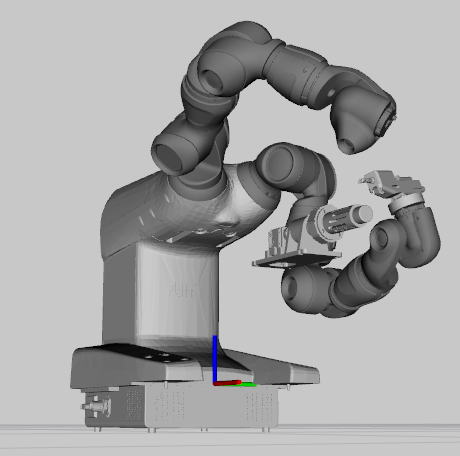}
    \caption{The simulation used to simulate the physics and visualize the robot movement.}
    \label{fig:yumi_simulation}
    % \vspace{-5mm}
\end{figure}

The simulation shown in Fig. \ref{fig:yumi_simulation} is the replicate setup of the real environment where the demonstrated trajectory will be applied after the optimization. The simulation provides collision detection in addition to visualization and inverse kinematic computation. For a given pose, the simulation computes the associated joint movement and detects potential collisions if the action was to be carried out. This information will be used to determine the optimal policy. 

The modelled trajectory is first used to create a 3D discretized space within which encloses all the demonstrated trajectories. The discretization is defined by a user-specified step size, $\delta$, and it represents the unit length for each grid in space. This creates a 3D space of size L, W, and H in x, y and z directions respectively, where $\{L, W, H\} \in \mathbb{R}^3$.  

The mean and variance of the trajectory are adopted to construct the constrained space for the optimization problem. Motion primitives with larger variance imply multiple possible trajectories and therefore the policy will aim to find the optimal path within the bounded space. On the other hand, the ones with smaller variance imply movement with higher precision, hence the policy should deviate less from the mean trajectory. By constraining the search area within the variance and applying a teacher's understanding of the subtrajectory to the learning policy, it can avoid the exploration of sub-optimal region and action and thus improving the learning efficiency. 

The goal of the RL agent is to find the optimal path such that it can complete a demonstrated task with minimal overall joint and end-effectors movements and without collision while ensuring the smoothness of the trajectory. To this end, the RL agent is trained under the guidance of the defined reward function as follows

\begin{align*}
r(\mathbf{s}, \mathbf{s}') = - \frac{\Delta\Phi}{7} - \frac{|\mathbf{s}'-\mathbf{s}|}{\left | (\mathbf{s}_{\text{mean}}'-\mathbf{s}_{\text{mean}}) \right |} - \angle({\mathbf{s}},{\mathbf{s}}') + 10
\end{align*}

The reward function consists of 4 terms. The first term is the average of $\Delta\Phi$ which is the sum of the 7 joint angle difference of the robot between the current state and the next state. The second term compares the pose length between the current state, $s$, and the next states, $s'$, to that of the mean trajectory, $|s_\text{mean}'-s_\text{mean}|$. The third term, $\angle({\mathbf{s}},{\mathbf{s}}')$, is the angle between two vectors, obtained from the previous state, the current state and the next state. \textcolor{black}{As the optimal trajectory is defined based on its overall trajectory length in both the joint space and the end-effector as well as the smoothness of the end-effector. The first two terms aim to penalize for action that resulted in higher joint changes and larger displacement between two consecutive states. The third term penalizes actions that resulted in sharp turn between states. The last term was determined heuristically to facilitate the convergence of the agent.}

The RL agent is represented by a deep neural network which takes the current state as the input and outputs the 126 possible actions. The state considered is only the translation term. The inclusion of orientation in the optimization will largely increase the number of possible states and the complexity of the policy and thus will be addressed in the future work. Each action corresponds to a point in space around the mean of the next point. 125 of them are uniformly sampled within the standard deviations of axes around the next mean point and the additional action represents the current state, i.e. stationary. For each step, the RL agent either takes a random valid action or the optimal action according the current policy. The rotation component is realized by finding the closest position on the mean trajectory to the next state and using the corresponding orientation as its rotation component. The location and the orientation are the pose which is used to calculate the inverse kinematics for the robot, before running in simulation for visualization and collision detection. Episode begins from the starting pose and ends at the terminal pose of the mean trajectory. The RL agent is trained at every step through mini-batch of samples from the experience buffer. When collision is detected, the current iteration is restarted.

After training, the optimal trajectory is generated by continuously taking the optimal action of the current state from the start point to the end of the mean trajectory. The corresponding joint trajectory is determined through inverse kinematics.

\section{EXPERIMENTS AND RESULTS}

\begin{figure*}[ht]
    \centering
    \includegraphics[width=18cm]{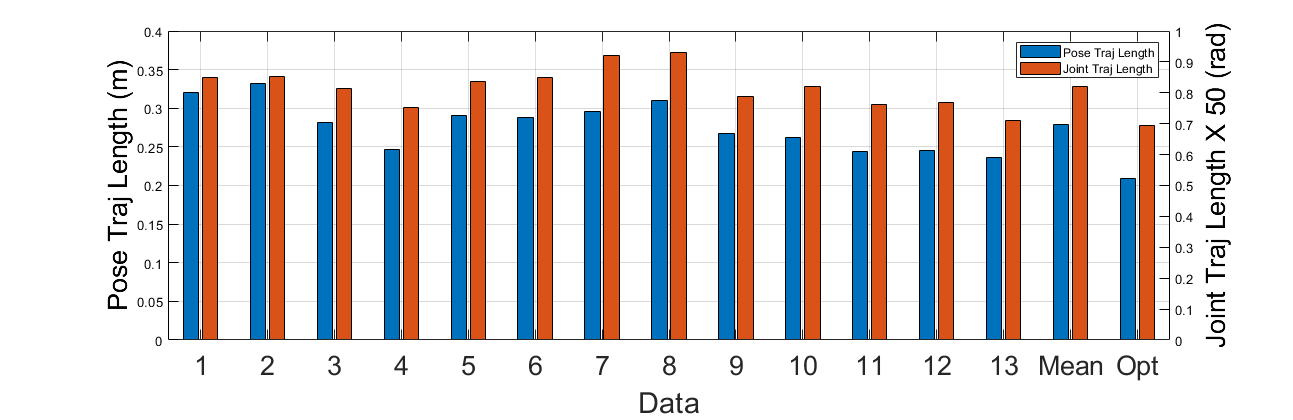}
    \caption{Comparisons of trajectory length among different human demonstrations and the optimal trajectory. \textcolor{black}{The first 10 data are the training data, used to form the bounded space, while the 11 to 13 are the best human demonstrations from the testing dataset used for trajectory comparisons. }}
    \label{fig:trajLenComp}
    % \vspace{-3mm}
\end{figure*}

\subsection{Experimental Setup}
To validate the proposed framework, we first collected a set of human demonstrated trajectories on single-handed stitching. During the demonstration, suturing was performed multiple times on the same location of the mandrel while the entire motion was recorded through stereo camera as described in the previous context. In total, 10 demonstrations were performed, each of which contained 6 DoF temporal poses of the mandrel and the suture device. A simple moving average smoothing with a window size of 10 was applied to smooth out sudden hand movements and/or vision misdetection. 

Demonstrations were first transformed to the mandrel frame of reference, and the DTW was performed to align temporally all the trajectories. The mean and variance of the trajectories were computed to construct a constrained space for exploration and exploitation during optimization. After which, the modelled mean trajectory was served as the basis for the optimization while the variance was used as the constraints for exploration.  

\begin{figure}[H]
% \vspace{-5mm}
    \centering{
    \subfloat[\scriptsize{}]
    {\includegraphics[width=0.85\linewidth]{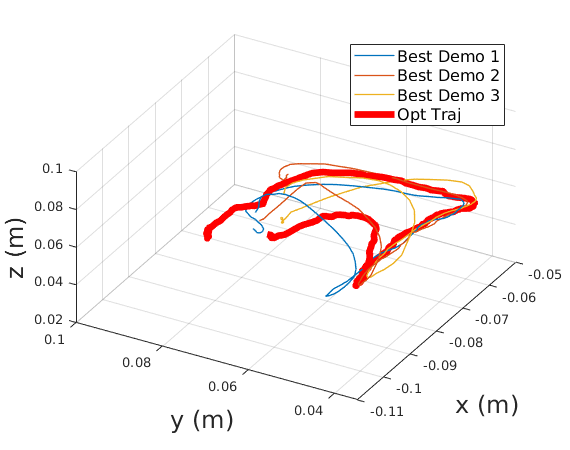}}}
    \newline
    \centering{
    \subfloat[\scriptsize{}]
    {\includegraphics[width=0.85\linewidth]{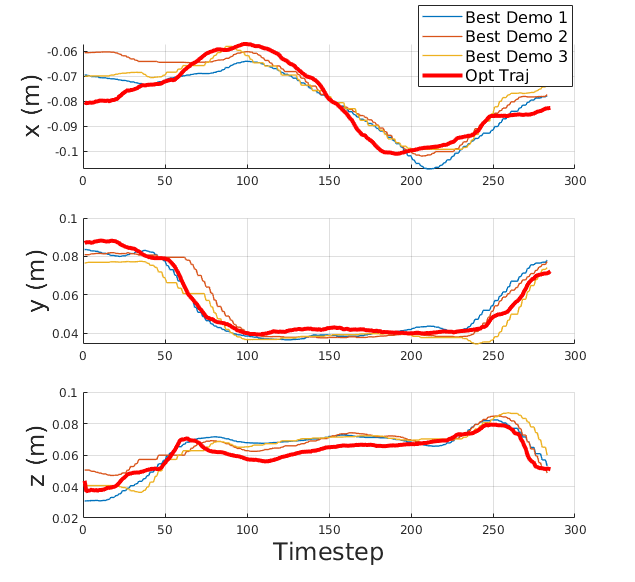}}}
    \caption{Comparison of the end-effector trajectories between the 3 best human demonstrations \textcolor{black}{from the testing dataset} and the optimized one. (a) shows the comparison in the Cartesian space and (b) compares temporally the differences in each DoF between the trajectories.}
    \label{fig:mean_opt_comp}
\end{figure} 

The optimization process was done using Gazebo simulation and visualizing using Rviz on the Robotics Operating System (ROS). The physical properties and location of different components in the simulation were mimicked to the real world setting. Gazebo was used to simulate the physical properties of the systems. In specific, it was mainly used to simulate the interaction between objects, i.e. collision. The robot used to carry out the single handed task was chosen as the ABB IRB14000 YuMi robot. For a given trajectory in the frame of its end-effectors, the inverse kinematic of the suturing hand was computed and the joint movement and objects interactions were visualized through Rviz.

During optimization, the RL training algorithm started from the mean trajectory, and the RL agent interacted with the simulation world to learn the policy at every step based on the reward function defined earlier. It identified the optimal path by exploring within the space where the variance was large while strictly following the mean of all the demonstrated trajectory when the variance was small. 

There were a few hyperparameters to be determined for tuning purpose during policy learning, in which the process was done mostly through trial and error and with some experience. This included, the learning rate, discount factor, mini-batch size, number of episodes and number of time-steps in one episode. The parameters of the neural network are summarized in Tab. \ref{tab:NN_param}. During policy learning, if a collision occurred, or the boundary condition, i.e. the variance, was met, the simulation would restart. The whole learning process stopped until convergence.  

The RL agent was represented by a deep neural network to map between states and actions. It consisted of 2 fully-connected hidden layers with 400 neurons in the first layer and 200 in the second, both with ReLU activation functions. The input to the agent was the current state, while the outputs were the Q-values of all possible actions. The update of the parameters in the neural network was done through mini-batch gradient fashion. A mini-batch of  training data was randomly selected from the experience buffer to update the network until convergence. The reward and the target Q-value was given and estimated based on its current state and the defined reward function. At the end of the process, the optimal path was generated by continuously taking the action with the highest Q-value until the end pose was reached. The start and the end pose were mainly given such that they corresponded to the first and the last poses of the mean trajectory.

\subsection{Results and Discussion}
\begin{table}[]
\centering
\caption{A summary of parameters used to train the RL agent.}
\begin{tabular}{ccc}
\multicolumn{1}{c}{\textbf{Parameter}} & \multicolumn{1}{c}{\textbf{Symbol}} & \multicolumn{1}{c}{\textbf{Values}} \\ \hline \hline
Step Size& $\delta$  & 0.002                                     \\\hline
Discount Factor& $\gamma$    & 0.99                               \\\hline
Episode& $Episode$   & 200                                \\\hline
Steps & $Step$ & 284 \\\hline
Exploration Rate& $\epsilon$   &$0.75 \frac{\text{Num\_Episode}- \text{Episode}}{\text{Num\_Episode}} $            \\\hline
Mini-Batch Size& $N$  & 32\\\hline
\end{tabular}
\label{tab:NN_param}
% \vspace{-3mm}
\end{table}
To validate the performance of the proposed framework, we performed quantitatively analyses on two spaces, one in the Cartesian space and the other in the joint space, in terms of the overall trajectory length and the smoothness. The target robot was the ABB IRB14000 YuMi robot and we utilized the Gazebo simulation to detect for collision during the execution. 

We compared the pose and joint trajectories of the optimal one \textcolor{black}{with all demonstrations from the training set and the three best human demonstrations from the testing dataset} as well as the mean trajectory. The results are shown in Fig. \ref{fig:trajLenComp}. The optimized trajectory showed the best results among all the demonstrations. The optimized had trajectory lengths of 0.21m in Pose and 796.98$\degree$ in the joint, while the best human demonstration had that of \textcolor{black}{0.24m} in Pose and \textcolor{black}{814.68$\degree$} in joint angle. 

In addition to the trajectory length, we also compared the smoothness of a pose trajectory. The smoothness of a pose trajectory was found by computing the mean of the change in the angle between three consecutive points along a trajectory. From the definition, the lower the changes, the smoother a trajectory is. We also compared the smoothness with the three best demonstrations. The results for the them were \textcolor{black}{27.19$\degree$, 27.56$\degree$ and 28.68$\degree$} respectively while the optimized one was 22.55$\degree$. 

The visual comparisons between the best demonstrations and the optimized trajectory are shown in Fig. \ref{fig:mean_opt_comp}. As clearly shown, the optimized trajectory not only appeared to have shorter trajectory, but also appeared to be smoother among the three. The quantitative analysis has validated the proposed framework is capable of learning an optimal trajectory from the sub-optimal trajectories. In the results, it has shown that the learned trajectory outperforms any of the demonstrated ones in terms of the trajectory length and the smoothness. The shorter length and the smoother trajectory in the optimized trajectory implied shorter and smoother execution of the learned task.

\section{CONCLUSIONS}
In summary, we proposed a framework that extends from the general LfD to teach a robot to perform a task using scarce and sub-optimal demonstrations. In this paper, we wanted a robot to learn a task from human demonstrated trajectories. Instead of directly learn the state-action mapping from them, the demonstrations were served as constraints to our optimization framework. During the training process, the RL agent interacted with the simulation environment. The training of the RL agent was guided by a carefully engineered reward function. In our experiments, we have shown that the proposed framework was able to find the optimal trajectory that outperformed any of the human demonstrated trajectory in terms of trajectory length and smoothness. In addition, the trajectory also took into account the collision when performed joint trajectory planning which would not be generally considered if the conventional mapping method was adopted to the learn a trajectory.

Using only a limited amount of demonstrations, not only we avoided the need to learn from scratch, but also shortened the amount of time needed for RL agent training by constraining the unexplored space. The current work can be served as a basis for a range of potential extensions. We validated the framework on a suturing task, which involves fine movement, but this could be easily extended to other domain. Besides, the robot adopted was the ABB YuMi robot and the proposed framework has shown its applicability to work on other types of robots. 

In this work, we used a discretized approach to perform the optimization and we only considered the positional information during training. Finer discretization and taking the orientation into consideration would result in longer training time. Therefore, in the future work, it is necessary to consider continuous RL approaches such as DDPG, using the orientation information for training and incorporating bimanual task execution.

\addtolength{\textheight}{-12cm}   % This command serves to balance the column lengths

%%%%%%%%%%%%%%%%%%%%%%%%%%%%%%%%%%%%%%%%%%%%%%%%%%%%%%%%%%%%%%%%%%%%%%%%%%%%%%%%

%%%%%%%%%%%%%%%%%%%%%%%%%%%%%%%%%%%%%%%%%%%%%%%%%%%%%%%%%%%%%%%%%%%%%%%%%%%%%%%%

%%%%%%%%%%%%%%%%%%%%%%%%%%%%%%%%%%%%%%%%%%%%%%%%%%%%%%%%%%%%%%%%%%%%%%%%%%%%%%%%

\bibliographystyle{IEEEtran}
\bibliography{ICRA20}

\end{document}